\documentclass{article}
\usepackage{spconf,amsmath,graphicx}
\usepackage{booktabs}
\usepackage{amsfonts}
\usepackage{color,soul}
\usepackage{multirow}
\usepackage[caption=false,font=footnotesize]{subfig}

\usepackage[hidelinks]{hyperref}
\usepackage{algorithm}
\usepackage[noend]{algpseudocode}
\usepackage{multirow}

\title{Unified Generator-Classifier for Efficient Zero-Shot Learning}

\name{
Ayyappa Kumar Pambala, Titir Dutta, Soma Biswas}
\address{Department of Electrical Engineering, Indian Institute of Science, Bangalore.}

\begin{document}
\ninept
\maketitle
\begin{abstract}

Generative models have achieved state-of-the-art performance for the zero-shot learning problem, but they require re-training the classifier every time a new object category is encountered.
The traditional semantic embedding approaches, though very elegant, usually do not perform at par with their generative counterparts.
In this work, we propose an unified framework termed {\em GenClass}, which integrates the generator with the classifier for efficient zero-shot learning, thus combining the representative power of the generative approaches and the elegance of the embedding approaches.
End-to-end training of the unified framework not only eliminates the requirement of additional classifier for new object categories as in the generative approaches, but also facilitates the generation of more discriminative and useful features.
Extensive evaluation on three standard zero-shot object classification datasets, namely AWA, CUB and SUN shows the effectiveness of the proposed approach.
The approach without any modification, also gives state-of-the-art performance for zero-shot action classification, thus showing its generalizability to other domains.
\end{abstract}


%

\section{Introduction}
\label{sec:intro}

New categories of objects are continuously being discovered, and this has motivated the zero-shot learning (ZSL) problem.
Unlike traditional supervised classification, where the object classes remain the same during both training and testing, in ZSL, the goal is to classify objects belonging to completely unseen classes, which are not present during training.
In real scenarios, usually we will not have any prior information whether the test objects are from seen or unseen classes, and this is termed as the generalized zero-shot learning (GZSL). 
For addressing this problem, we assume that some semantic information about both the seen and unseen classes are available, which is shared between them.
Although several approaches \cite{ALE}\cite{DEViSE}\cite{SJE}\cite{ESZSL}\cite{SAE} have been proposed, recognizing completely unseen objects using only their semantic representations is still a challenging task and needs to be addressed thoroughly~\cite{FGAN}\cite{VERMA}\cite{CVAE}.

The ZSL approaches in literature can broadly be divided into two categories. (1) The more traditional {\em embedding based approaches} try to find the mapping from visual space to semantic space~\cite{ALE}\cite{ESZSL}\cite{SAE}\cite{DEViSE}, or semantic space to visual space~\cite{Tri-Factorization}\cite{SAE}\cite{RelationNetwork}, or to a common intermediate space~\cite{SSE}; (2) Due to the enormous success of generative models like GAN~\cite{goodfellow2014generative}, VAE~\cite{kingma2013auto}, recent works~\cite{CVAE}\cite{FGAN}\cite{VERMA}~have used generative approaches which usually outperform the embedding based methods by a significant margin. 
Here, the generated features are used to train a classifier, thus transforming the ZSL task to a classical supervised classification problem.
Thus, the classifier needs to be trained each time a new class is encountered.

In this work, we propose a novel, unified framework termed {\em GenClass} by integrating the generator with the classifier. 
This generator-classifier module is trained adversarially with the discriminator which distinguishes the real/fake samples.
The generator aims to generate realistic samples using the attributes from both seen and unseen classes. 
The integrated classifier is designed to take a pair of image features~(one real and another fake) as input, and distinguish whether they belong to the same or different classes.
Since the real images are not available for unseen classes, we propose to utilize the generated image samples from unseen classes to incorporate the unseen class-discrimination ability into the classifier.
This helps in reducing the bias towards the seen classes.

Compared to the existing generative approaches which first try to generate realistic examples and then learn a classifier separately, end-to-end training of the unified framework has two significant advantages, namely (1) The generated features are more discriminative and tuned towards the final goal, thus we require very few generated examples for satisfactory performance; (2) There is no need of training a separate classifier which needs to be re-trained every time new classes are included. 
Extensive experiments on three ZSL datasets for object classification, namely AWA \cite{AWA}, CUB \cite{CUB2011} and SUN \cite{SUN} and also two datasets for action recognition, namely UCF101 \cite{SoomroUcf101} and HMDB51 \cite{KuehneHmdb} illustrate the effectiveness of the proposed approach.
Our contributions in this work are 
\begin{enumerate}
	\item We propose a novel unified framework which integrates the generative process with the classifier. The entire framework is trained end-to-end which results in improved performance and also eliminates the need for training a separate classifier. 
	\item The proposed approach is general and can potentially be used with different generative approaches.
	\item Extensive experimental evaluation for both object and action recognition in zero-shot setting shows the effectiveness of the proposed method as well as its generalizability for diverse domains.
\end{enumerate}

\section{Proposed Approach}
\label{sec:problem}

In this work, we propose a novel, unified generative framework termed {\em GenClass} which integrates the generative approach and a classifier for the task of ZSL.
First, we explain the different notations used and then describe the proposed framework in details. \\ \\
\textbf{Notations:} Let us consider that we have images and corresponding labels from a set of $K$ number of classes, $\mathcal{C}_{seen}=\{1,...,K\}$. 
As per the ZSL-protocol~\cite{GoodBadUgly}, we also have access to the attribute vectors $\mathbf{a}^{(s)} \in \mathbb{R}^{d_a}, s \in \mathcal{C}_{seen}$, which provides a unique higher level description of each class.
Thus the training data is given by $\mathcal{D}_{seen}=\{\mathbf{x}_i,\mathbf{y}_i,\mathbf{a}_i^{(s)}\}_{i=1}^{N}$.
Here, $\mathbf{x}_i \in \mathbb{R}^{d_x}$ is the feature representation of the $i^{th}$ image and
$\mathbf{y}_i \in \mathcal{C}_{seen}$ is its corresponding label-annotation.
$\mathbf{a}_i^{(s)} \in \mathbb{R}^{d_a}$ is the corresponding attribute vector representing that $\mathbf{x}_i$ belongs to category $(s \in \mathcal{C}_{seen})$.
For traditional ZSL, during testing, the images belong to the unseen categories only, and thus the goal is to classify the test image to one of the classes in the \emph{unseen} class set, $\mathcal{C}_{unseen}=\{K+1,...,K+L\}$, where $\mathcal{C}_{seen} \cap \mathcal{C}_{unseen}=\phi$.
In contrast, for GZSL, the test image can belong to either a seen or an unseen class, and so the test image has to be classified as one of $\mathcal{C}_{seen} \cup \mathcal{C}_{unseen}$.
Although no image data is available for the classes in $C_{unseen}$ while training, the class-attributes $\mathbf{A}_u = \{ \mathbf{a}^{(u)} | u \in \mathcal{C}_{unseen} \}$ are assumed to be available. \\
Existing generative approaches~\cite{FGAN}\cite{VERMA}\cite{CVAE} in ZSL generates pseudo (referred to as fake here) data for the unseen classes using their attributes.
This is augmented with the seen class training data and used for training a classifier which achieves impressive performance. 
\cite{FGAN} generates features using conditional GAN and uses a soft-max classifier, while~\cite{VERMA}\cite{CVAE} generates the features using conditional VAE and uses SVM to classify the augmented data.
In contrast, the main contribution of the proposed approach is that instead of generating image features and training a classifier separately, we integrate the classifier with the generator and the whole framework is trained in an end-to-end manner.
In this work, we have used the state-of-the-art approach based on GAN~\cite{FGAN} as the base network, though the proposed framework can work for other generative networks as well.
We first describe briefly the base network. 
\subsection{Background: GAN-based Base Network}
In the proposed work, the base generative network is based on Wasserstein Generative Adversarial Network~(WGAN) as in~\cite{FGAN}.
It consists of two modules - a generator~($G$) and a discriminator~($D$), which are trained in an adversarial fashion to capture the underlying distribution of the training data.
But unlike~\cite{FGAN}, we do not use the classifier that can classify the input into one of the seen classes. 
Instead, we have the integrated classifier which can not only handle the features from unseen classes, but can also be used seamlessly during testing. 
Using the training data of the seen classes $\mathcal{D}_{seen}$, the weights $\theta_{G}$ and $\theta_{D}$ of the generator~($G$) and discriminator~($D$) are learnt by optimizing the following min-max objective function
\begin{equation}
\theta_G^*, \theta_D^* = \min_{\theta_G} \max_{\theta_D} \mathcal{L}_{WGAN}
\end{equation}
where, the WGAN-loss function~\cite{ImprovedWGAN} is defined as,
\begin{equation}
\begin{aligned}
\mathcal{L}_{WGAN} =  \mathbb{E}[D(\mathbf{x}|\mathbf{a})] - \mathbb{E}[D(\tilde{\mathbf{x}}|\mathbf{a})] - \lambda(GP)
\end{aligned}
\end{equation}
Here, $\{\mathbf{x}, \mathbf{a}\} \in \mathcal{D}_{seen}$. 
We avoid the sample number and class indicators to avoid the notational clutter.
$\mathbf{\tilde{x}}$ is the fake sample generated by $G$, i.e. $\mathbf{\tilde{x}} = G(\mathbf{z}|\mathbf{a})$, where $\mathbf{z}$ is sampled from a pre-defined noise distribution $p_z$.
$GP$ is the gradient-penalty, enforced on the output of $\mathcal{D}$, which takes care of the instability issue~\cite{ImprovedWGAN} and is defined as:
$GP = \mathbb{E}[(\left \|  \nabla_{\hat{\mathbf{x}}} D(\hat{\mathbf{x}}|\mathbf{a})\right \|_2 - 1)^2], \text{ where } \hat{\mathbf{x}} = \alpha\mathbf{x} + (1-\alpha) \mathbf{\tilde{x}}, 
\text{ with } \alpha \sim U(0,1) $.
$\lambda$ is the penalty coefficient, and the suggested value~($\lambda=10$)~\cite{ImprovedWGAN} has been used in our work. 
\color{black}
\subsection{Proposed {\em GenClass}}
Here, we describe in details the proposed unified generator-classifier framework, {\em GenClass}.
We want to design the integrated classifier such that the following criteria are satisfied:
(1) It can generate discriminative features for the unseen classes, which are tuned for the application of GZSL; (2) It can be used during testing for both seen and unseen classes, thus eliminating the need for a separate classifier.
With this motivation, we design the integrated classifier, $C_I: \mathcal{X} \times \mathcal{X} \rightarrow \{0, 1\} $ which takes a pair of input features~(one fake and one real) and classifies the pair as belonging to the same or different classes.  
\begin{figure}[t!]
	\centering
	\includegraphics[width=\linewidth]{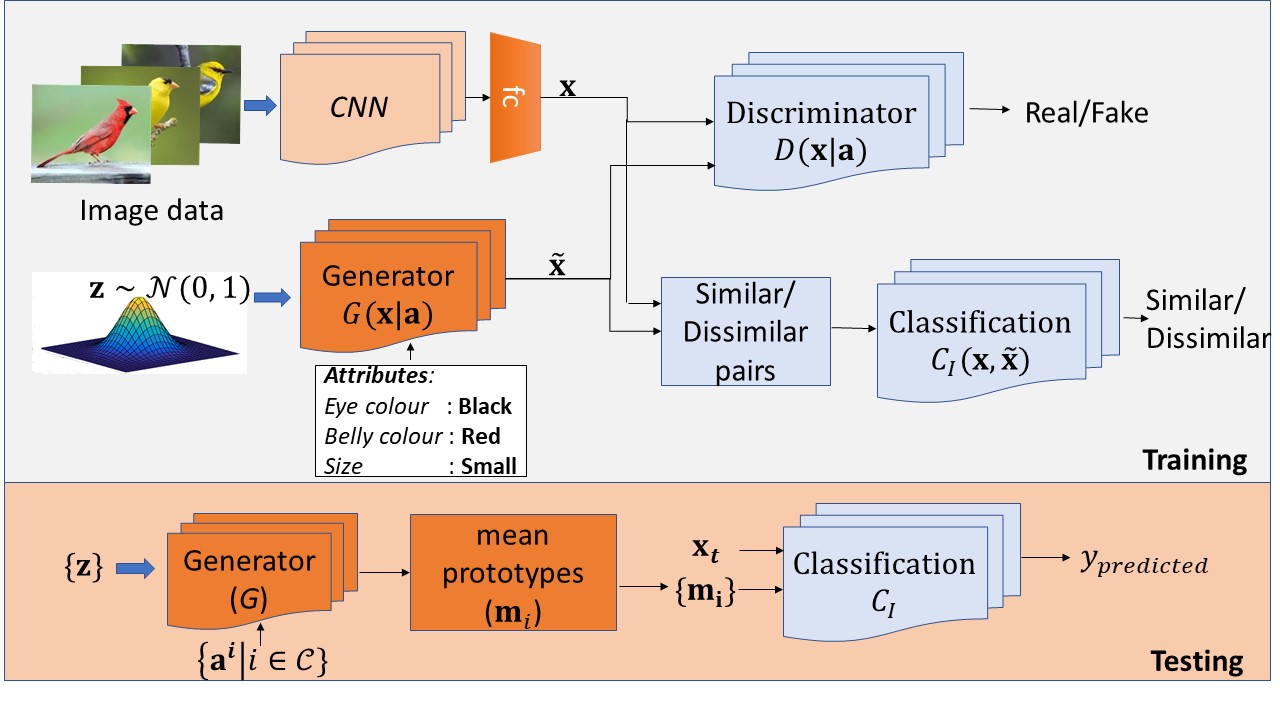}
	\caption{Illustration of the proposed \emph{GenClass} framework. No re-training of the classifier is required during testing.}
	\label{flowChart_zershot_action}
\end{figure}
To train $C_I$, we first generate a set of similar and dissimilar pairs from the seen data as
\begin{align}
\label{sim_seen}
\mathcal{D}_{seen}^{similar}=\{\mathbf{x}_i,\mathbf{\tilde{x}}_j | \text{ s.t., } y_i, y_j \in \mathcal{C}_{seen} \text{ }\& \text{ } y_i=y_j\}  \nonumber \\
\mathcal{D}_{seen}^{dissimilar}=\{\mathbf{x}_i,\mathbf{\tilde{x}_j} | \text{ s.t., } y_i, y_j \in \mathcal{C}_{seen} \text{ } \& \text{ } y_i \neq y_j \}
\end{align} 
$C_I$ takes $(\mathbf{x}_i,\mathbf{\tilde{x}}_j)$ as input and learns to predict $0$ for $(\mathbf{x}_i,\mathbf{\tilde{x}}_j) \in \mathcal{D}_{seen}^{dissimilar}$ and  $1$ for $(\mathbf{x}_i,\mathbf{\tilde{x}}_j) \in \mathcal{D}_{seen}^{similar}$, respectively by minimizing the following mean squared loss,
\begin{equation}
\mathcal{L}_{s} = \mathbb{E}_{(\mathbf{x}_i, \mathbf{\tilde{x}}_j) \sim (\mathcal{D}_{seen}^{similar} \cup \mathcal{D}_{seen}^{dissimilar}) } (C_I(\mathbf{x}_i, \mathbf{\tilde{x}}_j) - \mathbf{1}(y_i, y_j))^2
\end{equation}
where $\mathbf{1}(y_i,y_j) = 1$ if and only if $y_i=y_j$ and  $0$ otherwise. \\ \\
{\bf Generalization to unseen classes:}
The loss $\mathcal{L}_s$ enforces the generated features of the seen classes to be  discriminative. 
However, the learned weights in $C_I$ will still be biased towards the seen classes.
To reduce this bias and to improve generalizability, we further use the generated features from the attributes for classes in $\mathcal{C}_{unseen}$ to modify the weights.
Specifically, we incorporate an additional loss to learn the weights $\theta_{C_I}$,
\begin{equation}
\mathcal{L}_{u} = \mathbb{E}_{(\mathbf{\tilde{x}}_i, \mathbf{\tilde{x}}_j) \sim (\mathcal{D}_{unseen}^{similar} \cup \mathcal{D}_{unseen}^{dissimilar}) }(C_I(\mathbf{\tilde{x}}_i, \mathbf{\tilde{x}}_j) - \mathbf{1}(y_i, y_j))^2
\end{equation} where $\mathbf{\tilde{x}}_i = G(\mathbf{z}|\mathbf{a}^{(u)})$, such that $\mathbf{z}_i \sim p_z$ and $\mathbf{a}^{(u)} \in \mathbf{A}_u$.
Since no real examples of the unseen classes are available, the similar and dissimilar pairs in this case are generated based on only the fake samples as
\begin{align}
\label{sim_unseen}
\mathcal{D}_{unseen}^{similar} = \{\mathbf{\tilde{x}}_i,\mathbf{\tilde{x}}_j | \text{ s.t., } y_i, y_j \in \mathcal{C}_{unseen} \text{ } \& \text{ } y_i = y_j\} \nonumber \\
\mathcal{D}_{unseen}^{dissimilar} = \{\mathbf{\tilde{x}}_i,\mathbf{\tilde{x}}_j | \text{ s.t., } y_i, y_j \in \mathcal{C}_{unseen} \text{ } \& \text{ } y_i \neq y_j\}
\end{align}
Finally, the overall loss function for the proposed $C_I$ module is,
\begin{equation}
\mathcal{L}_{C_I} = \mathcal{L}_{s} + \gamma \mathcal{L}_{u}
\end{equation}
where $\gamma$ is a hyper-parameter weighting the different components of the loss.
Thus, the overall loss function of the unified framework {\em GenClass} can be expressed as,
\begin{equation}
\mathcal{L}_{GC} = \mathcal{L}_{WGAN} + \mathcal{L}_{C_I}
\end{equation}
The final optimized parameters of \emph{GenClass} are learned as,
\begin{equation}
	\theta_G^*, \theta_D^*, {\theta^*_{C_I}}= \min_{\theta_G, \theta_{C_I}} \max_{\theta_{D}}  \mathcal{L}_{GC}
\end{equation}

\begin{algorithm}[t!]
	\caption{Algorithm for training {\em GenClass}} 
	\label{algo1}
	\begin{algorithmic}[1]
		\State \textbf{Input}: $\mathcal{D}_{seen}, \mathbf{A}_u$.
		\State \textbf{Initialize}: Randomly initialize the parameters of the discriminator ($\theta_D$), generator ($\theta_G$) and integrated classifier ($\theta_{C_I}$).
		\State \textbf{Requirement}: Learning rate $\beta$, number of iterations of discriminator per generator iteration $ = n_{dis}$, batch size $B$, $p_z \sim \mathcal{N}(0, 1)$.
		\For{number of training iterations} 
		
		\For{$i=1,2,....,n_{dis}$}   
		\State Sample $\{\textbf{x}_i, \textbf{a}_i\}_{i=1}^B \sim D_{seen}$, $\{\textbf{z}_i\}_{i=1}^B$ $\sim p_z$. 
		\State $g_{\theta_D}\leftarrow \nabla_{\theta_D}  \frac{1}{B} \sum\limits_{i=1}^{B}{\big[D\big( G(\textbf{z}_i|\textbf{a}_i)\big)  - D(\textbf{x}_i|\textbf{a}_i) + \lambda (GP_i) \big]} $
		\State $\theta_D \leftarrow \theta_D - \beta \text{Adam}(\theta_{D},g_{\theta_D}) $
		\EndFor	
		
		\State Sample 
		\begin{itemize}
		\item $\mathcal{D}_{seen}^{batch}=\{\textbf{x}_i, \textbf{a}_i\, \mathbf{z}_i\}_{i=1}^B, \{ \mathbf{x}_i, \mathbf{a}_i \} \in \mathcal{D}_{seen}, \mathbf{z}_i \sim p_z$.
		\item $\mathcal{D}_{unseen}^{batch}=\{ \mathbf{a}_i, \mathbf{z}_i \}_{i=1}^B, \mathbf{a}_i \in \mathbf{A}_u, \mathbf{z}_i \sim p_z$.
		\end{itemize}
		\State 
		Generate fake data for seen and unseen classes
		
		 $\tilde{\textbf{x}_i} = G(\textbf{z}_i|\textbf{a}_i), \mathbf{a}_i \in \mathcal{D}_{seen}^{batch} \cup \mathcal{D}_{unseen}^{batch}$.
		\State Generate batch-wise similar and dissimilar pairs for both 
		
		$\mathcal{D}_{seen}^{batch}$ and $\mathcal{D}_{unseen}^{batch}$ as in \eqref{sim_seen} and \eqref{sim_unseen}, respectively.
		\State $g_{\theta_{C_I}} \leftarrow \nabla_{\theta_{C_I}}\mathcal{L}_{C_I}$
		
		\State $g_{\theta_G} \leftarrow  \nabla_{\theta_G} \frac{1}{B}\sum\limits_{i=1}^{B}{\big[\mathcal{L}_{C_I} - D(G(\textbf{z}_i|\textbf{a}_i)) \big]}$
		
		\State $\theta_{C_I} \leftarrow \theta_{C_I} - \beta \text{Adam}(\theta_{C_I},g_{\theta_{C_I}})$
		\State $\theta_{G} \leftarrow \theta_{G} - \beta \text{Adam}(\theta_{G},g_{\theta_{G}}) $
		\EndFor
		\State \textbf{Output} : $\theta_{C_I}^{*}, \theta_D^{*}, \theta_G^{*}$
	\end{algorithmic}
\end{algorithm}

The end-to-end training of the entire framework using this loss function facilitates the generator to generate more relevant and discriminative features.
The algorithm used for training {\em GenClass} is given in Algorithm~\ref{algo1}.
Fig.~\ref{flowChart_zershot_action} depicts an illustration of the same.  \\ \\
\textbf{Testing:} In addition to facilitating generation of more discriminative features, the integrated classifier in {\em GenClass} serves as a classifier during testing, thereby eliminating the need of an additional classifier.

Given the attribute of each test class, we generate $n_g$ number of samples ($n_g \approx 50$).
Using these generated samples, we compute the mean representative of each class.
The mean class representatives effectively serve as the prototypes for the respective test classes.
If $\mathbf{m}_i$ denotes the representative for the $i^{th}$ class~($i \in \mathcal{C}, \text{ where, }\mathcal{C}=\mathcal{C}_{unseen}$ for ZSL and $\mathcal{C}=\mathcal{C}_{seen} \cup \mathcal{C}_{unseen}$ for GZSL), 
\begin{equation}
	\mathbf{m}_i = \frac{1}{n_g}\sum_{k=1}^{n_g} G(\mathbf{z}_k|\mathbf{a}^{(i)})
\end{equation}
The incoming test query is paired with all $\mathbf{m}_i$'s and passed through $C_I$ to obtain the similarity prediction. 
The final class-prediction for the test data ($\mathbf{x_t}$) is obtained as
\begin{equation}
y_{predicted} = \underset{1 \leq i \leq |\mathcal{C}|}{\mathrm{arg max}} ~ C_I(\mathbf{m}_i, \mathbf{x_t}) 
\end{equation}
We observe in our experiments that mean vectors generated from few generated samples is sufficient for satisfactory performance.\\ 
\textbf{Implementation Details:} In our implementation, $G$ and $D$ have MLP architecture with two hidden layers consisting of $4096$ units each. 
The activation functions at the hidden layers and the final layer of the generator are LeakyReLU and ReLU respectively.
$C_I$ has a single hidden layer with $1024$ nodes with LeakyReLU activation and the output is a single node activated with sigmoid function. 
The noise $\mathbf{z}$ is sampled from a Gaussian distribution $p_z=\mathcal{N}(0,1)$.
The value of the hyper-parameter $\gamma~(0 \leq \gamma \leq 1)$ is set empirically for individual dataset. 
For optimization, we use Adam optimizer with learning rate $10^{-4}$ and batch size of $64$ across all the experiments.


\section{Experimental Evaluation}
Here, we describe the extensive experiments performed to evaluate the usefulness of proposed approach for the task of zero-shot object and action classification. 
\subsection{Zero-Shot Object Classification} 
We first describe the experiments for zero-shot object classification. \\ 
\textbf{Datasets Used:} We evaluate our performance on two fine-grained datasets, namely CUB~\cite{CUB2011} and SUN~\cite{SUN} and one coarse-grained dataset, namely AWA~\cite{AWA}. CUB contains 11,788 images from 200 classes, annotated with 312 attributes. SUN contains 14,340 images of 717 classes, annotated with 102 attributes. AWA contains 30,475 images of 50 classes, annotated with 85 attributes. 
We evaluate the proposed approach for both the standard split (SS) and proposed split (PS) given by \cite{GoodBadUgly}. 
For all three datasets, we use the features extracted using Resnet101~\cite{Resnet}, released by \cite{GoodBadUgly}.  
For the semantic class embedding, we use the manually labeled attributes of each class for AWA (85-d), CUB (312-d) and SUN (102-d).\\
\begin{table*}[th!]
\footnotesize
	\resizebox{\textwidth}{!}{%
		\begin{tabular}{|l|cccccc|ccccccccc|}
			\hline
			& \multicolumn{6}{c|}{Zero-Shot Learning} & \multicolumn{9}{c|}{Generalized Zero-Shot Learning} \\ \hline
			& \multicolumn{2}{c}{CUB} & \multicolumn{2}{c}{SUN} & \multicolumn{2}{c|}{AWA} & \multicolumn{3}{c}{CUB} & \multicolumn{3}{c}{SUN} & \multicolumn{3}{c|}{AWA} \\ \hline
			Method & SS & PS & SS & PS & SS & PS & U & S & H & U & S & H & U & S & H \\ \hline
			DEVISE~\cite{DEViSE} & 53.2 & 52.0 & 57.5 & 56.5 & 72.9 & 54.2 & 23.8 & 53.0 & 32.8 & 16.9 & 27.4 & 20.9 & 13.4 & 68.7 & 22.4 \\
			SJE~\cite{SJE} & 55.3 & 53.9 & 57.1 & 53.7 & 76.7 & 65.6 & 23.5 & 59.2 & 33.6 & 14.7 & 30.5 & 19.8 & 11.3 &\textbf{74.6} & 19.6 \\
			LATEM~\cite{LATEM} & 49.4 & 49.3 & 56.9 & 55.3 & 74.8 & 55.1 & 15.2 & 57.3 & 24.0 & 14.7 & 28.8 & 19.5 & 7.3 & 71.7 & 13.3 \\
			ESZSL~\cite{ESZSL} & 55.1 & 53.9 & 57.3 & 54.5 & 74.7 & 58.2 & 12.6 &\textbf{63.8} & 21.0 & 11.0 & 27.9 & 15.8 & 6.6 & 75.6 & 12.1 \\
			ALE~\cite{ALE} & 53.2 & 54.9 & 59.1 & 58.1 & 78.6 & 59.9 & 23.7 & 62.8 & 34.4 & 21.8 & 33.1 & 26.3 & 16.8 & 76.1 & 27.5 \\
			DCN \cite{DCN}             & -      & 56.2 & - &61.8 &- & 65.2& 28.4 &60.7 &  38.7    &  25.5  & 37.0   &30.2      &25.5    & 84.2 & 39.1  \\			 \hline
			CVAE~\cite{CVAE} & - & 52.1 & - & 61.7 & - &\textbf{71.4} & - & - & 34.5 & - & - & 26.7 & - & - & 47.2 \\
			SE~\cite{VERMA} & 60.3 & 59.6 & 64.5 & 63.4 &\textbf{83.8} & 69.5 & 41.5 & 53.3 & 46.7 & 40.9 & 30.5 & 34.9 & 56.3 & 67.8 &61.5 \\
			f-CLSWGAN~\cite{FGAN} & - & 57.3 & - & 60.8 & - & 68.2 & 43.7 & 57.7 & 49.7 & 42.6 &\textbf{36.6} &39.4 & \textbf{57.9} & 61.4 & 59.6 \\ \hline
			\textbf{GenClass} &\textbf{64.0} &\textbf{60.5} & \textbf{66.6} & \textbf{63.5} & 83.5 & 70.1 & \textbf{46.1} & 61.2 & \textbf{52.5} & \textbf{48.0} & 34.8 & \textbf{40.3} &  55.4 & 70.6 & \textbf{62.1} \\ \hline
		\end{tabular}%
	}
	\caption{Evaluation of {\em GenClass} using average per-class top-1 accuracy (\%) and comparison with the state-of-the-art. ``SS" and ``PS" denote standard and proposed split~\cite{GoodBadUgly} for  ZSL. ``H" denotes the harmonic mean of the performance for unseen ``U" and seen ``S" data for GZSL.}
\label{objectresults_new}
\end{table*}
The results of the proposed approach for the three datasets for both the standard ZSL and the GZSL scenarios are shown in Table~\ref{objectresults_new}.
We compare the proposed approach with the state-of-the-art embedding based approaches DEViSE~\cite{DEViSE}, ALE~\cite{ALE}, CMT~\cite{CMT}, CONSE~\cite{CONSE}, ESZSL~\cite{ESZSL}, DCN~\cite{DCN} and also the state-of-the-art generative approaches CVAE~\cite{CVAE}, f-CLSWGAN~\cite{FGAN}, SE~\cite{VERMA}.
The results of all the other approaches are taken directly from \cite{GoodBadUgly}.
We observe that for the standard ZSL scenario, the proposed approach performs significantly better than all the embedding based approaches.
For CUB and SUN datasets, it performs better than the generative approaches and it's close to the best for AWA dataset.
For GZSL, on all datasets, the proposed approach obtains the best performance in terms of the H factor.
We observe that for the GZSL, proposed approach performs very well for the unseen categories, but the performance is slightly less than the state-of-the-art for seen categories, but overall (H factor) it still outperforms all the other approaches.
\subsection{ZSL for Action Classification:} We now evaluate the proposed approach without any modification to another very important problem, i.e. zero-shot action classification.
\textbf{Datasets Used:}  For this experiment, we have used UCF101~\cite{SoomroUcf101}, which is a challenging database collected from YouTube containing 13,320 action videos from 101 classes.
We consider randomly selected 51 classes as seen classes and the remaining 50 are taken as unseen~\cite{Mishragenerative}.
HMDB51~\cite{KuehneHmdb} is another realistic database which contains a total of 6,766 video clips from 51 action classes, each having at least 101 clips. 
As in ~\cite{Mishragenerative}, we use randomly selected 26 classes as seen and the remaining as unseen.
For each dataset, we repeat experiment on 30 independent train/test splits~\cite{Mishragenerative} and the mean and standard deviations are reported for a fair comparison.\\
\begin{table}[h!]
	\centering
\footnotesize
	\begin{tabular}{|l|c|cc|}
		\hline
		& {A} & \multicolumn{2}{c|}{W} \\ \hline
		Method & UCF101  & UCF101 & HMDB51 \\ \hline
		HAA  \cite{HAA} &  14.9 $\pm$ 0.8    &  N/A & N/A \\
		DAP \cite{lampert2014attribute} & 14.3 $\pm$ 1.3& N/A & N/A \\
		IAP \cite{lampert2014attribute} & 12.8 $\pm$ 2.0& N/A & N/A \\
		ST \cite{ST} & N/A & 13.0 $\pm$ 2.7 & 10.9 $\pm$ 1.5 \\
		SJE \cite{SJE} & 12.0 $\pm$ 1.2  & 9.9 $\pm$ 1.4 & 13.3 $\pm$ 2.4 \\
		UDA \cite{UDA}& 13.2 $\pm$ 1.9  & N/A & N/A \\
		Bi-Dir \cite{WangZero} & 20.5 $\pm$ 0.5   & 18.9 $\pm$ 0.4 & 18.6 $\pm$ 0.7 \\
		ESZSL \cite{ESZSL} & N/A & 15.0 $\pm$ 1.3  & 18.5 $\pm$ 2.0 \\ 
		GGM \cite{Mishragenerative} & 22.7 $\pm$ 1.2   & 17.3 $\pm$ 1.1 & 19.3 $\pm$ 2.1 \\ \hline
		\textbf{GenClass} &  \textbf{27.5 $\pm$ 3.4}& \textbf{19.8 $\pm$ 2.0}   &  \textbf{21.3 $\pm$ 3.6} \\ \hline
	\end{tabular}
\caption{Evaluation of {\em GenClass} for zero shot action recognition. ``A" and ``W" denote human annotated attributes and word2vec-embedding, respectively. Manual attributes for HMDB51 dataset is not available. Average per-class top-1 accuracy is reported.}
\label{actionresults}
\end{table}
The results are summarized in Table~\ref{actionresults}.
Here, we compare the results with the state-of-the-art approaches for the ZSL setting.
The results of all the other approaches are taken directly from~\cite{Mishragenerative}.
We observe that the proposed approach performs significantly better as compared to the state-of-the-art for both the action datasets in standard ZSL setting.
This shows the generalizability of the approach to other domains. 
\section{Analysis of the Proposed Approach}
In this section, we analyze the effect of different number of generated samples, effect of the different losses, quality of generated samples, etc. on the proposed framework. \\ 
\textbf{Effect of number of generated samples: }
In the proposed approach, since the classifier is trained along with the generator in an end-to-end manner, the generator learns to generate samples which are more useful for the task of classification.
Thus, the number of samples required to be generated is significantly less as compared to the case when the generator and classifier are trained separately.
We increase the number of generated samples and observed that using proposed framework, performance almost saturates after $20-50$ samples for ZSL on all datasets. 
For example, even with a single generated sample, \emph{GenClass} achieves an accuracy of $56.83$ on CUB, which increases to $60.5$ with $20$-samples and saturates after that. 
In contrast, standard generative-model based approaches~\cite{FGAN}\cite{CVAE} require few hundreds of samples to achieve reasonable performance.
We find similar pattern for GZSL as well. \\ 
\textbf{Quality of generated samples:} Another advantage of the end-to-end training of the generator and the classifier is that the generated samples are much more discriminative.
We test this hypothesis by performing a simple experiment on the CUB dataset, in which we use the same training process as mentioned in previous section. 
But instead of using the integrated classifier for the final classification, we use the softmax classifier (as in~\cite{FGAN})  trained using the generated samples. 
\begin{table}[h!]
	\centering
\begin{tabular}{|l|c|ccl|}
	\hline
	\multirow{2}{*}{Method} & \multirow{2}{*}{ZSL}  & \multicolumn{3}{c|}{GZSL} \\ 
	\cline{3-5}
	 &   & \multicolumn{1}{c|}{U} & \multicolumn{1}{c|}{S} & H \\ \hline
	f-CLSWGAN \cite{FGAN} & 57.3 & 43.7 & 57.7 & 49.7 \\ \hline
	\textbf{GenClass}+soft-max & \textbf{58.4} & \textbf{44.7} & \textbf{59.6} & \textbf{51.1} \\ \hline
\end{tabular}
\caption{Performance of softmax classifier with features generated by the base network~\cite{FGAN} and those generated using {\em GenClass} for both ZSL and GZSL scenarios.}
\label{quality-comparision}
\end{table}
We observe from Table \ref{quality-comparision} that the softmax classifier is able to obtain higher performance with the features generated using {\em GenClass} as compared to the original disjoint framework. 
This justifies that the generated features using {\em GenClass} are more discriminative. \\ 
\textbf{Effect on other generative approaches:}
To test the generalizability of the integrated classifier module, we have conducted experiment by replacing the base generative model~(WGAN in our framework) with CVAE~(as in~\cite{CVAE}). We summarize the performance of proposed approach with this modification in Table~\ref{genclass-vae}. 
We observe that for both ZSL and GZSL-settings, the integrated framework yields better or comparable accuracy.
\begin{table}[h!]
	\footnotesize
	\centering
	\begin{tabular}{|l|ccc|ccc|}
		\hline
		& \multicolumn{3}{c|}{ZSL} & \multicolumn{3}{c|}{GZSL} \\ \hline
		Method & \multicolumn{1}{c|}{CUB} & \multicolumn{1}{c|}{SUN} & AWA & \multicolumn{1}{c|}{CUB} & \multicolumn{1}{c|}{SUN} & AWA \\ \hline
		CVAE~\cite{CVAE} & 52.1 & 61.7 & 71.4 & 34.5 & 26.7 & 47.2 \\ \hline
		CVAE + $C_I$ & \textbf{57.9}  &\textbf{62.2} & 68.5 & \textbf{39.4} & \textbf{29.0} & \textbf{49.8} \\ \hline
	\end{tabular}
\label{genclass-vae}
\caption{Performance of integrated classifier module of \emph{GenClass} with CVAE~\cite{CVAE} as base network. Average per-class top-1 accuracies and harmonic mean are reported for ZSL and GZSL respectively. }
\end{table}
\section{Conclusion}
\label{sec:majhead}
In this paper, we proposed a novel, unified framework {\em GenClass} which integrates the generator with a classifier for the task of ZSL and GZSL.
The proposed framework eliminates the need of a separate classifier for the generative approaches and also ensures that the generated features are more discriminative.
Extensive experiments performed on different zero-shot object classification and action classification datasets show the effectiveness of the proposed approach as well as its generalizability for different domains.

\clearpage
\bibliographystyle{IEEEbib}
\small
\bibliography{strings,refs}

\end{document}